﻿
\documentclass[sigconf,screen]{acmart}
\usepackage{stfloats}
\usepackage{algorithm}
\usepackage{algorithmic}
\usepackage{textcomp}
\usepackage{hyperref}

\AtBeginDocument{%
  \providecommand\BibTeX{{%
    \normalfont B\kern-0.5em{\scshape i\kern-0.25em b}\kern-0.8em\TeX}}}

\copyrightyear{2020}
\acmYear{2020}
\setcopyright{acmcopyright}
\acmConference[MM '20]{Proceedings of the 28th ACM International Conference on Multimedia}{October 12--16, 2020}{Seattle, WA, USA}
\acmBooktitle{Proceedings of the 28th ACM International Conference on Multimedia (MM '20), October 12--16, 2020, Seattle, WA, USA}
\acmPrice{15.00}
\acmDOI{10.1145/3394171.3416284}
\acmISBN{978-1-4503-7988-5/20/10}

\begin{document}
\fancyhead{}

\title{Video Relation Detection with Trajectory-aware Multi-modal Features}

\author{Wentao Xie}
\authornote{Work done during internship at YITU Technology.}
\authornote{Both authors contributed equally to this research.}
\affiliation{
  \institution{Beihang University}
}
\email{wynter_xie@outlook.com}

\author{Guanghui Ren}
\authornotemark[2]
\affiliation{
  \institution{YITU Technology}
  }
\email{sundrops.ren@gmail.com}

\author{Si Liu}
\authornote{Corresponding author.}
\affiliation{
  \institution{Beihang University}
  }
\email{liusi@buaa.edu.cn}




\begin{abstract}
  Video relation detection problem refers to the detection of the relationship between different objects in videos, such as spatial relationship and action relationship. In this paper, we present video relation detection with trajectory-aware multi-modal features to solve this task.
  Considering the complexity of doing visual relation detection in videos, we decompose this task into three sub-tasks: object detection, trajectory proposal and relation prediction. We use the state-of-the-art object detection method to ensure the accuracy of object trajectory detection and multi-modal feature representation to help the prediction of relation between objects. Our method won the first place on the video relation detection task of Video Relation Understanding Grand Challenge in ACM Multimedia 2020 with 11.74\% mAP,
 which surpasses other methods by a large margin.
\end{abstract}



\keywords{Video Relation detection;Object trajectory detection;Relation prediction}


\maketitle

\section{Introduction}
Video relation detection is to find all object trajectories and relation between them as triplet 	
\textlangle{} subject,predicate,object \textrangle{}
 in a video. It bridges visual information and linguistic information, enabling the cross-modal information transformation. Comparing to other computer vision tasks like object detection and semantic segmentation, visual relation detection requires not only localizing and categorizing single object but also understanding the interaction between different objects. To capture the relation between objects, more information of the video content need to be utilized.

 Visual relation detection in videos is much more difficult than in static images. On the one hand, spatial-temporal localization of objects is needed instead of just spatial localization. This requires to track the same object in different frames along the temporal axis. On the other hand, relations between objects become more variable. Relations between the same object pair can change during time and some temporal-related relations will be introduced, which makes it more difficult to predict the relations. Thus, it is hard to directly apply existing methods on visual relation detection\cite{liao2020ppdm} or scene graph generation\cite{ren2020scene} to this task.

 Several methods have been proposed to solve the problem. \cite{shang2017video} firstly introduced a baseline for video relation detection. It firstly divides the video into several segments with same temporal length using sliding window. Secondly it performs video relation detection in segments by a object trajectory proposal module and a relation prediction module. Finally it generates the video relation detection results in the video by merging the result of those segments greedily. \cite{tsai2019video} introduced a Gated Spatio-Temporal Energy Graph model as the relation prediction module of the baseline proposed by \cite{shang2017video}. By constructing a Conditional Random Field on a fully-connected spatio-temporal graph, the statistical dependency between relational entities spatially and temporally can be better exploited. \cite{qian2019video} introduced graph convolutional network into the relation predictor, which takes better advantages of the spatial-temporal context.

\begin{figure*}[ht]
    \centering
    \includegraphics[width=1\textwidth]{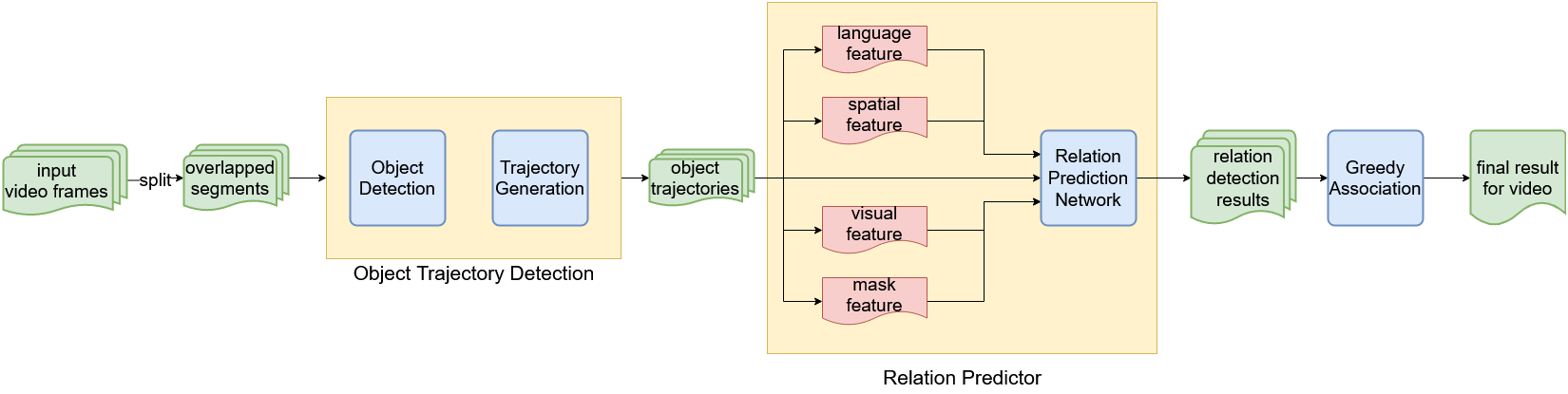}
    \caption{Framework of our method}
    \label{framework}
\end{figure*}

 In this paper, we propose a method for video relation detection problem. We follow the scheme of \cite{shang2017video} to build our system with a object trajectory detector module and a relation predictor module. For object trajectory detector, we first perform object detection for each video frame with the state-of-the-art detector Cascade R-CNN\cite{cai2018cascade} with ResNeSt101\cite{Zhang2020ResNeStSN} as backbone. Then we use a Dynamic Programming Algorithm improved from seq-NMS\cite{han2016seq-nms} to associate the object detection results of all frames and generate trajectories for each object. For relation predictor, we combine motion feature, visual feature, language feature and location mask feature for each trajectory pair to predict the relation between them. The use of multi-modal feature helps to increase the accuracy of relation prediction. The framework of our method is shown in Figure.~\ref{framework} Our method achieved the first place on the video relation detection task of Video Relation Understanding Grand Challenge\cite{shang2019relation} in ACM Multimedia 2020.

\section{Object Trajectory Detection}

\subsection{Object Detection}
We choose Cascade R-CNN as our object detection model and ResNeSt101 as the network backbone. To train the object detector, we extract frames from each video to build the training set and validation set. Due to the high similarity between frames in the same video, using all frames is not necessary. Thus, we sample at most 15 key frames, whose bounding boxes are drawn by human, for each video in the training set of VidOR dataset\cite{shang2019annotating}. The training set of detection consists of 97221 images extracted by the above method. Also, the validation set consists of 31220 human-labeled frames extracted from the validation set of VidOR dataset.

During training, we notice that the class imbalance issue exists in our training procedure. Classes with more annotations(adult, child, baby, etc.) have high AP up to 0.7 while classes with less annotations(crocodile, frisbee, etc.) get low AP close to zero. To overcome this imbalance issue, we extend our training set with part of the images from MS COCO dataset, which is more balanced than our training set.

During testing, we perform object detection for all video frames and keep bounding boxes that have confidence score higher than 0.01 as our final detection results.

\subsection{Trajectory Generation}
We take the tracking-by-detection strategy to generate object trajectories. Based on the object detection results of all video frames, we use a Dynamic Programming algorithm improved from seq-NMS to associate bounding boxes that belong to the same object and generate the trajectory. This algorithm consists of two part: Graph Building and Trajectory Selection. By regarding each bounding box as a node of the graph, we can link the bounding boxes that are likely to belong to the same object and from consecutive frames. After that, paths of the graph represent trajectories and we can run Dynamic Programming algorithm to pick paths that are more likely to be a object trajectory.

\textbf{Graph Building: } First, we regard each bounding box as a node and build the initial graph with no edge between them. Let $category_{t,n}$,$conf_{t,n}$,$bbox_{t,n}$,$in_{t,n}$,$out_{t,n}$ represent the object category, confidence score, bounding box, set of precursor nodes and set of successor nodes of the $n_{th}$ bounding box in frame $t$. We set $in$ and $out$ to empty for all nodes initially. Then, for each $t(0<=t<T$, T is the frame count$)$ and all possible $(i,j)$ that satisfy $node_{t,i}$ and $node_{t+1,j}$ exist, if $category_{t,i}$ and $category_{t+1,j}$ are the same and the IoU of $bbox_{t,i}$ and $bbox_{t+1,j}$ is higher than a threshold, add $node_{t+1,j}$ to $out_{t,i}$ and add $node_{t,i}$ to $in_{t+1,j}$. By doing this, we link bounding boxes pair from consecutive frames that has a IoU higher than the threshold. We set the threshold to 0.2 in our experiments.

\begin{figure*}[ht]
    \centering
    \includegraphics[width=0.8\textwidth]{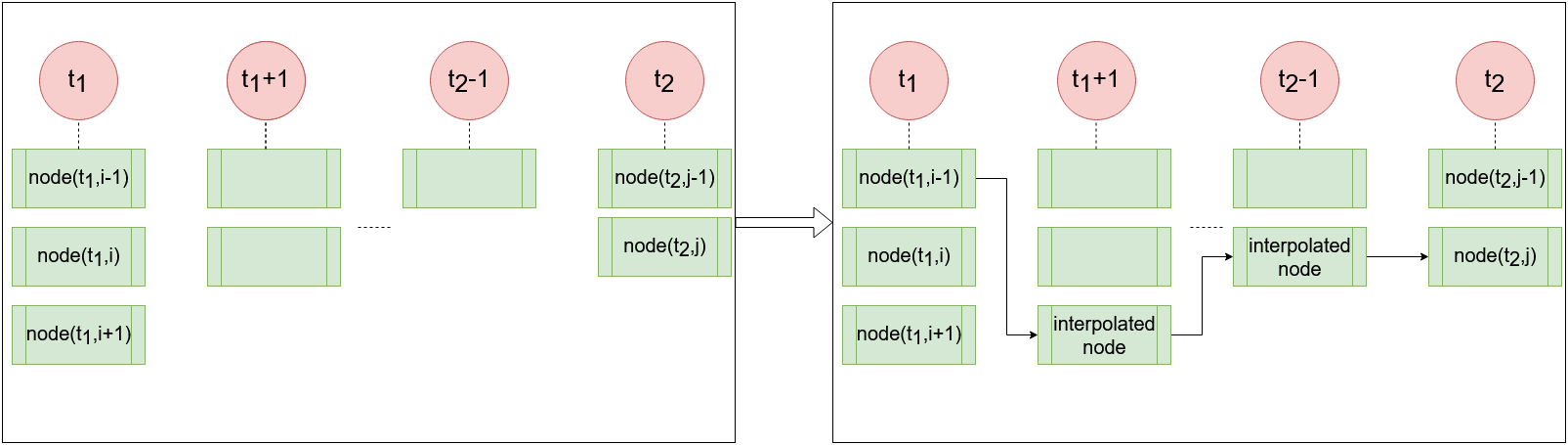}
    \caption{Cross-frame Linking Mechanism}
    \label{cflm}
\end{figure*}

We notice that when the camera or object in the video is moving violently, the IoU of bounding boxes that belong to the same object from consecutive frames will be very low. In this case, the original seq-NMS algorithm won't link them, causing the lost of tracking. To solve this problem, we introduce a new linking mechanism. First,for bounding boxes $B1=(x_1,y_1,w_1,h_1)$ and $B2=(x_2,y_2,w_2,h_2)$, we define scale\_ratio and area\_ratio as:
\begin{equation}
  scale\_ratio(B1,B2) =
  \begin{cases}
    \frac{h_2w_1}{w_2h_1}& \frac{h_2}{w_2}>\frac{h_1}{w_1}\\
    \frac{h_1w_2}{w_1h_2}& \frac{h_2}{w_2}<=\frac{h_1}{w_1}
  \end{cases}
\end{equation}
\begin{equation}
  area\_ratio(B1,B2) =
  \begin{cases}
    \frac{w_1h_1}{w_2h_2}& w_1h_1>w_2h_2\\
    \frac{w_2h_2}{w_1h_1}& w_1h_1<=w_2h_2
  \end{cases}
\end{equation}
Then, for each $(t_1,t_2,i,j)$ that satisfies
\begin{itemize}
  \item [1)]
  $0<=t_1<t_2<=T,t_2 - t_1 > 1$
  \item [2)]
  $node_{t_1,i}$, $node_{t_2,j}$ exist
  \item [3)]
  $scale\_ratio(bbox_{t_1,i},bbox_{t_2,j}) > 0.5$
  \item [4)]
  $area\_ratio(bbox_{t_1,i},bbox_{t_2,j}) > 0.5$
\end{itemize}
we create a path from $node_{t_1,i}$ to $node_{t_2,j}$ by interpolating nodes in each time $t(t_1<t<t_2)$, as shown in Fig.~\ref{cflm}. The $bbox$ of the interpolated node is obtained by linear interpolation of $bbox_{t_1,i}$ and $bbox_{t_2,j}$. The confidence score of the interpolated node will be set to 0. By applying this linking mechanism, the trajectory generation module is more robust to violent movement of camera and objects. In our experiments, we limit $t_2 - t_1$ to be less than 8 to make a trade-off between performance and complexity.

\textbf{Trajectory Selection: } After building the graph, we can regard a full path(path that can not be extended) of the graph as a object trajectory and take sum of confidence score of nodes in the path as the score of the path. Then, we repeatedly select
path with the highest score and remove the nodes of the path from the graph. We achieve this by Dynamic Programming Algorithm used in \cite{han2016seq-nms}. Trajectories selected by the algorithm will be returned as the trajectory detection result.

\section{Relation Prediction}

\begin{figure}[hb]
    \centering
    \includegraphics[width=0.8\linewidth]{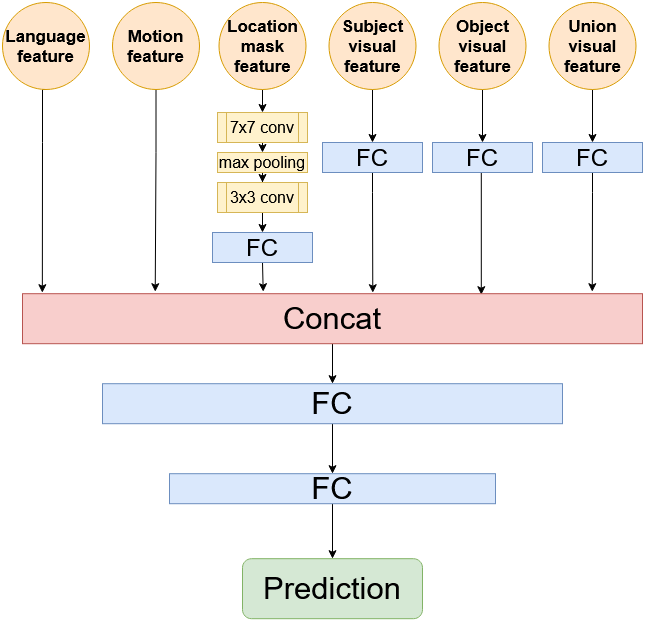}
    \caption{Relation Prediction Network}
    \label{rp}
\end{figure}

Follow the scheme of \cite{shang2017video}, we first divide the video into overlapped segments with same length and perform object trajectory detection in all segments. We set segment length to 32 frames and overlap length to 16 frames in our experiments. After that, we predict the relation between all possible object pairs in the same segment.

\subsection{Features}
To fully capture the video context and temporal movement, we use multi-modal features, including motion feature, visual feature, language feature and location mask feature, to help the relation prediction.

\textbf{Motion Feature: }
For a trajectory pair in 32 frame segment, we first calculate the location feature following method used in \cite{sun2019video} for frame 0, 8, 16, 24 and 31. Let $feat_t$ be the feature calculate for frame $t$. To capture the relative location of the pair in the static frame, we generate static feature $feat_{static}$ by concatenating all the features calculated for frame 0, 8, 16, 24 and 31. To capture the dynamic movement of the pair, we generate dynamic feature $feat_{dynamic}$ by concatenating $feat_8-feat_0$, $feat_{16}-feat_0$, $feat_{24}-feat_0$, $feat_{31}-feat_0$.

\textbf{Visual Feature: }
Due to the high complexity of extracting feature from video using network like I3D\cite{carreira2017quo}, we choose to only extract visual feature from static frame using 2-D network. Most previous work used the object detection model to extract feature for relation prediction. However, detection model focus on the category of single object in the image. It can not capture the relation information properly. Thus, we use a scene graph generation model\cite{tang2020unbiased} pre-trained on Visual Genome Dataset\cite{krishna2017visual} to extract feature for relation prediction to help better capturing the interaction between objects. We only use the the middle frame of the segment to extract feature. For each pair, we extract a 4096-d feature for bounding box of the subject, bounding box of the object and the union of their bounding box respectively.

\textbf{Language Feature: }
For language context, we follow \cite{sun2019video} to generate a 300-d feature for subject and object category respectively and concatenate them as the final language feature.

\textbf{Location Mask Feature: }
Since coordinates only have very limited ability in representing location, we further introduce the binary mask of the bounding box to better capture the relative location of subject and object. We follow the method of \cite{zellers2018neural} to generate a mask base on the bounding boxes of the subject and object in the middle frame of the segment as a input of the relation predictor.

\subsection{Network Design}
Using the features mentioned above as input, we design a simple neural network to predict the relation. The structure of the network is shown in Figure.~\ref{rp}.

After analysing the dataset, we find that about 99\% of object pairs in the training set have no more than one spatial relation and one action relation. Thus, we convert the multi-label classification problem appeared in VidOR\cite{shang2019annotating} Dataset to two single-label classification problem. We use focal loss\cite{lin2017focal} to supervise the spatial label and the action label separately to deal with the severe imbalance issue.

\section{Experiment}
In this section, we present experiment results in VidOR Dataset. We use the official evaluation code of the grand challenge to evaluate out results. More detail can be found in
\url{https://videorelation.nextcenter.org/mm20-gdc/task1.html}.

\begin{table*}[ht]
  \caption{Our detailed evaluation scores on VidOR test set (\%)}
  \label{tab:detail}
  \begin{tabular}{ccccccc}
    \toprule
    Method &mAP &R@50 &R@100 &tagging P@1 &tagging P@5 &tagging P@10\\
    \midrule
    Ours(colab-buaa) & 11.74 & 10.02 & 12.69 & 71.36 & 56.30 & 44.59\\
    \bottomrule
  \end{tabular}
\end{table*}

\subsection{Component Analysis}
\textbf{Object Trajectory Detection: } We adopt Cascade R-CNN with ResNeSt101 as our object detector and Dynamic Programming algorithm with cross-frame linking mechanism as trajectory generator. To prove the effectiveness of our trajectory generation algorithm, we firstly evaluate it in the optional task Video Object Detection. Since we don't submit our result for the optional task, we only compare our result on the validation set of VidOR Dataset with the result of the first place of the optional task in 2020 on the test set of VidOR Dataset. Table.~\ref{tab:op} shows that we surpass the first place of the optional task in 2020 by a large margin. Secondly, we compare the video visual relation detection results using dynamic programming with and without cross-frame linking mechanism. The results are shown in Table.~\ref{tab:traj}. CFLM means cross-frame linking mechanism. We can find that cross-frame linking mechanism increases the mAP from 8.84\% to 9.93\%.

\begin{table}[h]
  \caption{Comparison with state-of-the-art methods on the optinal task Video Object Detection  (\%)}
  \label{tab:op}
  \begin{tabular}{cc}
    \toprule
    Method  &mAP\\
    \midrule
    DeepBlueAI(on test set) & 9.66\\
    ours(on validation set) & \textbf{14.59}\\
    \bottomrule
  \end{tabular}
\end{table}

\begin{table}[h]
  \caption{Results using different trajectory generation methods on VidOR validation set  (\%)}
  \label{tab:traj}
  \begin{tabular}{cccc}
    \toprule
    Method &tagging P@1 &R@50 &mAP\\
    \midrule
    ours w/o CFLM &66.59 &8.30 & 8.84\\
    ours & \textbf{67.43} &\textbf{9.12} & \textbf{9.93}\\
    \bottomrule
  \end{tabular}
\end{table}

\textbf{Relation Prediction: } We use multi-modal features to proceed relation prediction. We perform 4 experiments using our multi-modal features without language feature, motion feature, visual feature and location mask feature respectively. As shown in Table.~\ref{tab:feat}, our method using all features outperforms other experiments, which shows the effectiveness of our multi-modal features. We can also find that using less feature doesn't decrease the mAP as much as not using cross-frame linking mechanism. It means that for current dataset and methods in video relation detection, robust trajectory detection matters more.

\begin{table}[h]
  \caption{Results using different features on VidOR validation set  (\%)}
  \label{tab:feat}
  \begin{tabular}{cccc}
    \toprule
    Method &tagging P@1 &R@50 &mAP\\
    \midrule
    Ours w/o language &66.70 &8.98 & 9.66\\
    Ours w/o motion &67.18 &9.01 & 9.86\\
    Ours w/o visual &65.99 &8.83 & 9.50\\
    Ours w/o mask &65.75 &9.08 & 9.74\\
    Ours & \textbf{67.43} &\textbf{9.12} & \textbf{9.93}\\
    \bottomrule
  \end{tabular}
\end{table}

\subsection{Comparison with state-of-the-art}
We compare our results with other methods in VidOR validation dataset. As shown in Table.~\ref{tab:sota}, our method outperform other methods by a large margin, which proves the effectiveness of our method.

\begin{table}[h]
  \caption{Comparison with state-of-the-art methods on VidOR validation set (\%)}
  \label{tab:sota}
  \begin{tabular}{cccc}
    \toprule
    Method &tagging P@1 &R@50 &mAP\\
    \midrule
    RELAbuilder\cite{zheng2019relation} &33.05 &1.58 & 1.47\\
    MAGUS.Gamma\cite{sun2019video} &51.20 &6.89 & 6.56\\
    VRD-STGC\cite{Liu_2020_CVPR} &48.92 &8.21 & 6.85\\
    Ours & \textbf{67.43} &\textbf{9.12} & \textbf{9.93}\\
    \bottomrule
  \end{tabular}
\end{table}

We use model ensemble strategy to further improve our result for the challenge task. Tabel.~\ref{tab:test} shows the comparison of our method and other methods in VidOR test dataset. We also outperform all other methods by a large margin.

\begin{table}[ht]
  \caption{Comparison with state-of-the-art methods on VidOR test set (\%)}
  \label{tab:test}
  \begin{tabular}{cc}
    \toprule
    Method &mAP\\
    \midrule
    RELAbuilder\cite{zheng2019relation} & 0.55\\
    MAGUS.Gamma\cite{sun2019video} & 6.31\\
    DeepBlueAI & 0.24\\
    GKBU & 3.28\\
    Zixuan Su & 5.99\\
    ETRI\_DGRC & 6.65\\
    Ours(colab-buaa) & \textbf{11.74}\\
    \bottomrule
  \end{tabular}
\end{table}

The detailed evaluation scores of our method on VidOR test set is shown in Table.~\ref{tab:detail}.

\section{Conclusion}

In this paper, we propose trajectory-aware multi-modal features for video relation detection. Finally, we achieved 11.74\% mAP, ranking the first place on the video relation detection task of Video Relation Understanding Grand Challenge in ACM Multimedia 2020.

\section{ACKNOWLEDGEMENTS}

This work was partially supported by the National Natural Science Foundation of China (Grant 61876177), Beijing Natural Science Foundation (Grant 4202034), Fundamental Research Funds for the Central Universities and Zhejiang Lab (No. 2019KD0AB04).

\newpage


\bibliographystyle{ACM-Reference-Format}
\bibliography{sample-base}


\begin{thebibliography}{18}


\ifx \showCODEN    \undefined \def \showCODEN     #1{\unskip}     \fi
\ifx \showDOI      \undefined \def \showDOI       #1{#1}\fi
\ifx \showISBNx    \undefined \def \showISBNx     #1{\unskip}     \fi
\ifx \showISBNxiii \undefined \def \showISBNxiii  #1{\unskip}     \fi
\ifx \showISSN     \undefined \def \showISSN      #1{\unskip}     \fi
\ifx \showLCCN     \undefined \def \showLCCN      #1{\unskip}     \fi
\ifx \shownote     \undefined \def \shownote      #1{#1}          \fi
\ifx \showarticletitle \undefined \def \showarticletitle #1{#1}   \fi
\ifx \showURL      \undefined \def \showURL       {\relax}        \fi
\providecommand\bibfield[2]{#2}
\providecommand\bibinfo[2]{#2}
\providecommand\natexlab[1]{#1}
\providecommand\showeprint[2][]{arXiv:#2}

\bibitem[\protect\citeauthoryear{Cai and Vasconcelos}{Cai and
  Vasconcelos}{2018}]%
        {cai2018cascade}
\bibfield{author}{\bibinfo{person}{Zhaowei Cai} {and} \bibinfo{person}{Nuno
  Vasconcelos}.} \bibinfo{year}{2018}\natexlab{}.
\newblock \showarticletitle{Cascade R-CNN: Delving Into High Quality Object
  Detection}.
\newblock  (\bibinfo{year}{2018}), \bibinfo{pages}{6154--6162}.
\newblock


\bibitem[\protect\citeauthoryear{Carreira and Zisserman}{Carreira and
  Zisserman}{2017}]%
        {carreira2017quo}
\bibfield{author}{\bibinfo{person}{Joao Carreira} {and} \bibinfo{person}{Andrew
  Zisserman}.} \bibinfo{year}{2017}\natexlab{}.
\newblock \showarticletitle{Quo Vadis, Action Recognition? A New Model and the
  Kinetics Dataset}.
\newblock  (\bibinfo{year}{2017}), \bibinfo{pages}{4724--4733}.
\newblock


\bibitem[\protect\citeauthoryear{Han, Khorrami, Paine, Ramachandran,
  Babaeizadeh, Shi, Li, Yan, and Huang}{Han et~al\mbox{.}}{2016}]%
        {han2016seq-nms}
\bibfield{author}{\bibinfo{person}{Wei Han}, \bibinfo{person}{Pooya Khorrami},
  \bibinfo{person}{Tom~Le Paine}, \bibinfo{person}{Prajit Ramachandran},
  \bibinfo{person}{Mohammad Babaeizadeh}, \bibinfo{person}{Honghui Shi},
  \bibinfo{person}{Jianan Li}, \bibinfo{person}{Shuicheng Yan}, {and}
  \bibinfo{person}{Thomas~S Huang}.} \bibinfo{year}{2016}\natexlab{}.
\newblock \showarticletitle{Seq-NMS for Video Object Detection.}
\newblock \bibinfo{journal}{\emph{arXiv: Computer Vision and Pattern
  Recognition}} (\bibinfo{year}{2016}).
\newblock


\bibitem[\protect\citeauthoryear{Krishna, Zhu, Groth, Johnson, Hata, Kravitz,
  Chen, Kalantidis, Li, Shamma, et~al\mbox{.}}{Krishna et~al\mbox{.}}{2017}]%
        {krishna2017visual}
\bibfield{author}{\bibinfo{person}{Ranjay Krishna}, \bibinfo{person}{Yuke Zhu},
  \bibinfo{person}{Oliver Groth}, \bibinfo{person}{Justin Johnson},
  \bibinfo{person}{Kenji Hata}, \bibinfo{person}{Joshua Kravitz},
  \bibinfo{person}{Stephanie Chen}, \bibinfo{person}{Yannis Kalantidis},
  \bibinfo{person}{Lijia Li}, \bibinfo{person}{David~A Shamma},
  {et~al\mbox{.}}} \bibinfo{year}{2017}\natexlab{}.
\newblock \showarticletitle{Visual Genome: Connecting Language and Vision Using
  Crowdsourced Dense Image Annotations}.
\newblock \bibinfo{journal}{\emph{International Journal of Computer Vision}}
  \bibinfo{volume}{123}, \bibinfo{number}{1} (\bibinfo{year}{2017}),
  \bibinfo{pages}{32--73}.
\newblock


\bibitem[\protect\citeauthoryear{Liao, Liu, Wang, Chen, Qian, and Feng}{Liao
  et~al\mbox{.}}{2020}]%
        {liao2020ppdm}
\bibfield{author}{\bibinfo{person}{Yue Liao}, \bibinfo{person}{Si Liu},
  \bibinfo{person}{Fei Wang}, \bibinfo{person}{Yanjie Chen},
  \bibinfo{person}{Chen Qian}, {and} \bibinfo{person}{Jiashi Feng}.}
  \bibinfo{year}{2020}\natexlab{}.
\newblock \showarticletitle{Ppdm: Parallel point detection and matching for
  real-time human-object interaction detection}. In
  \bibinfo{booktitle}{\emph{Proceedings of the IEEE/CVF Conference on Computer
  Vision and Pattern Recognition}}. \bibinfo{pages}{482--490}.
\newblock


\bibitem[\protect\citeauthoryear{Lin, Goyal, Girshick, He, and Doll{\'a}r}{Lin
  et~al\mbox{.}}{2017}]%
        {lin2017focal}
\bibfield{author}{\bibinfo{person}{Tsung-Yi Lin}, \bibinfo{person}{Priya
  Goyal}, \bibinfo{person}{Ross Girshick}, \bibinfo{person}{Kaiming He}, {and}
  \bibinfo{person}{Piotr Doll{\'a}r}.} \bibinfo{year}{2017}\natexlab{}.
\newblock \showarticletitle{Focal loss for dense object detection}. In
  \bibinfo{booktitle}{\emph{Proceedings of the IEEE international conference on
  computer vision}}. \bibinfo{pages}{2980--2988}.
\newblock


\bibitem[\protect\citeauthoryear{Liu, Jin, Xu, Gong, and Mu}{Liu
  et~al\mbox{.}}{2020}]%
        {Liu_2020_CVPR}
\bibfield{author}{\bibinfo{person}{Chenchen Liu}, \bibinfo{person}{Yang Jin},
  \bibinfo{person}{Kehan Xu}, \bibinfo{person}{Guoqiang Gong}, {and}
  \bibinfo{person}{Yadong Mu}.} \bibinfo{year}{2020}\natexlab{}.
\newblock \showarticletitle{Beyond Short-Term Snippet: Video Relation Detection
  With Spatio-Temporal Global Context}. In \bibinfo{booktitle}{\emph{IEEE/CVF
  Conference on Computer Vision and Pattern Recognition (CVPR)}}.
\newblock


\bibitem[\protect\citeauthoryear{Qian, Zhuang, Li, Xiao, Pu, and Xiao}{Qian
  et~al\mbox{.}}{2019}]%
        {qian2019video}
\bibfield{author}{\bibinfo{person}{Xufeng Qian}, \bibinfo{person}{Yueting
  Zhuang}, \bibinfo{person}{Yimeng Li}, \bibinfo{person}{Shaoning Xiao},
  \bibinfo{person}{Shiliang Pu}, {and} \bibinfo{person}{Jun Xiao}.}
  \bibinfo{year}{2019}\natexlab{}.
\newblock \showarticletitle{Video Relation Detection with Spatio-Temporal
  Graph}.
\newblock  (\bibinfo{year}{2019}), \bibinfo{pages}{84--93}.
\newblock


\bibitem[\protect\citeauthoryear{Ren, Ren, Liao, Liu, Li, Han, and Yan}{Ren
  et~al\mbox{.}}{2020}]%
        {ren2020scene}
\bibfield{author}{\bibinfo{person}{Guanghui Ren}, \bibinfo{person}{Lejian Ren},
  \bibinfo{person}{Yue Liao}, \bibinfo{person}{Si Liu}, \bibinfo{person}{Bo
  Li}, \bibinfo{person}{Jizhong Han}, {and} \bibinfo{person}{Shuicheng Yan}.}
  \bibinfo{year}{2020}\natexlab{}.
\newblock \showarticletitle{Scene Graph Generation With Hierarchical Context}.
\newblock \bibinfo{journal}{\emph{IEEE Transactions on Neural Networks and
  Learning Systems}} (\bibinfo{year}{2020}).
\newblock


\bibitem[\protect\citeauthoryear{Shang, Di, Xiao, Cao, Yang, and Chua}{Shang
  et~al\mbox{.}}{2019a}]%
        {shang2019annotating}
\bibfield{author}{\bibinfo{person}{Xindi Shang}, \bibinfo{person}{Donglin Di},
  \bibinfo{person}{Junbin Xiao}, \bibinfo{person}{Yu Cao}, \bibinfo{person}{Xun
  Yang}, {and} \bibinfo{person}{Tatseng Chua}.}
  \bibinfo{year}{2019}\natexlab{a}.
\newblock \showarticletitle{Annotating Objects and Relations in User-Generated
  Videos}.
\newblock  (\bibinfo{year}{2019}), \bibinfo{pages}{279--287}.
\newblock


\bibitem[\protect\citeauthoryear{Shang, Ren, Guo, Zhang, and Chua}{Shang
  et~al\mbox{.}}{2017}]%
        {shang2017video}
\bibfield{author}{\bibinfo{person}{Xindi Shang}, \bibinfo{person}{Tongwei Ren},
  \bibinfo{person}{Jingfan Guo}, \bibinfo{person}{Hanwang Zhang}, {and}
  \bibinfo{person}{Tatseng Chua}.} \bibinfo{year}{2017}\natexlab{}.
\newblock \showarticletitle{Video Visual Relation Detection}.
\newblock  (\bibinfo{year}{2017}), \bibinfo{pages}{1300--1308}.
\newblock


\bibitem[\protect\citeauthoryear{Shang, Xiao, Di, and Chua}{Shang
  et~al\mbox{.}}{2019b}]%
        {shang2019relation}
\bibfield{author}{\bibinfo{person}{Xindi Shang}, \bibinfo{person}{Junbin Xiao},
  \bibinfo{person}{Donglin Di}, {and} \bibinfo{person}{Tat-Seng Chua}.}
  \bibinfo{year}{2019}\natexlab{b}.
\newblock \showarticletitle{Relation Understanding in Videos: A Grand Challenge
  Overview}. In \bibinfo{booktitle}{\emph{Proceedings of the 27th ACM
  International Conference on Multimedia}}. \bibinfo{pages}{2652--2656}.
\newblock


\bibitem[\protect\citeauthoryear{Sun, Ren, Zi, and Wu}{Sun
  et~al\mbox{.}}{2019}]%
        {sun2019video}
\bibfield{author}{\bibinfo{person}{Xu Sun}, \bibinfo{person}{Tongwei Ren},
  \bibinfo{person}{Yuan Zi}, {and} \bibinfo{person}{Gangshan Wu}.}
  \bibinfo{year}{2019}\natexlab{}.
\newblock \showarticletitle{Video Visual Relation Detection via Multi-modal
  Feature Fusion}.
\newblock  (\bibinfo{year}{2019}), \bibinfo{pages}{2657--2661}.
\newblock


\bibitem[\protect\citeauthoryear{Tang, Niu, Huang, Shi, and Zhang}{Tang
  et~al\mbox{.}}{2020}]%
        {tang2020unbiased}
\bibfield{author}{\bibinfo{person}{Kaihua Tang}, \bibinfo{person}{Yulei Niu},
  \bibinfo{person}{Jianqiang Huang}, \bibinfo{person}{Jiaxin Shi}, {and}
  \bibinfo{person}{Hanwang Zhang}.} \bibinfo{year}{2020}\natexlab{}.
\newblock \showarticletitle{Unbiased Scene Graph Generation from Biased
  Training}.
\newblock \bibinfo{journal}{\emph{arXiv: Computer Vision and Pattern
  Recognition}} (\bibinfo{year}{2020}).
\newblock


\bibitem[\protect\citeauthoryear{Tsai, Divvala, Morency, Salakhutdinov, and
  Farhadi}{Tsai et~al\mbox{.}}{2019}]%
        {tsai2019video}
\bibfield{author}{\bibinfo{person}{Yaohung~Hubert Tsai},
  \bibinfo{person}{Santosh~K Divvala}, \bibinfo{person}{Louisphilippe Morency},
  \bibinfo{person}{Ruslan Salakhutdinov}, {and} \bibinfo{person}{Ali Farhadi}.}
  \bibinfo{year}{2019}\natexlab{}.
\newblock \showarticletitle{Video Relationship Reasoning Using Gated
  Spatio-Temporal Energy Graph}.
\newblock  (\bibinfo{year}{2019}), \bibinfo{pages}{10424--10433}.
\newblock


\bibitem[\protect\citeauthoryear{Zellers, Yatskar, Thomson, and Choi}{Zellers
  et~al\mbox{.}}{2018}]%
        {zellers2018neural}
\bibfield{author}{\bibinfo{person}{Rowan Zellers}, \bibinfo{person}{Mark
  Yatskar}, \bibinfo{person}{Sam Thomson}, {and} \bibinfo{person}{Yejin Choi}.}
  \bibinfo{year}{2018}\natexlab{}.
\newblock \showarticletitle{Neural Motifs: Scene Graph Parsing with Global
  Context}.
\newblock  (\bibinfo{year}{2018}), \bibinfo{pages}{5831--5840}.
\newblock


\bibitem[\protect\citeauthoryear{Zhang, Wu, Zhang, Zhu, Zhang, Lin, e~Sun, He,
  Mueller, Manmatha, Li, and Smola}{Zhang et~al\mbox{.}}{2020}]%
        {Zhang2020ResNeStSN}
\bibfield{author}{\bibinfo{person}{Hang Zhang}, \bibinfo{person}{Chongruo Wu},
  \bibinfo{person}{Zhongyue Zhang}, \bibinfo{person}{Yi Zhu},
  \bibinfo{person}{Zhi-Li Zhang}, \bibinfo{person}{Haibin Lin},
  \bibinfo{person}{Yu e Sun}, \bibinfo{person}{Tong He}, \bibinfo{person}{Jonas
  Mueller}, \bibinfo{person}{R. Manmatha}, \bibinfo{person}{Mengnan Li}, {and}
  \bibinfo{person}{Alexander~J. Smola}.} \bibinfo{year}{2020}\natexlab{}.
\newblock \showarticletitle{ResNeSt: Split-Attention Networks}.
\newblock \bibinfo{journal}{\emph{ArXiv}}  \bibinfo{volume}{abs/2004.08955}
  (\bibinfo{year}{2020}).
\newblock


\bibitem[\protect\citeauthoryear{Zheng, Chen, Chen, and Jin}{Zheng
  et~al\mbox{.}}{2019}]%
        {zheng2019relation}
\bibfield{author}{\bibinfo{person}{Sipeng Zheng}, \bibinfo{person}{Xiangyu
  Chen}, \bibinfo{person}{Shizhe Chen}, {and} \bibinfo{person}{Qin Jin}.}
  \bibinfo{year}{2019}\natexlab{}.
\newblock \showarticletitle{Relation Understanding in Videos}.
\newblock  (\bibinfo{year}{2019}), \bibinfo{pages}{2662--2666}.
\newblock


\end{thebibliography}

\end{document}